\documentclass[10pt,a4paper]{article}

\usepackage[margin=1in]{geometry}
\usepackage[T1]{fontenc}
\usepackage[utf8]{inputenc}
\usepackage{newtxtext,newtxmath}
\usepackage{microtype}

\usepackage{amsmath}
\usepackage{graphicx}
\usepackage{float}
\usepackage{booktabs}
\usepackage{tabularx}
\usepackage{array}
\usepackage[font=small,labelfont=bf]{caption}
\usepackage{placeins}
\usepackage[round,authoryear]{natbib}
\usepackage[colorlinks=true,citecolor=blue,linkcolor=blue,urlcolor=blue]{hyperref}

\makeatletter
\renewcommand\normalsize{%
  \@setfontsize\normalsize{10pt}{11pt}%
  \abovedisplayskip 10pt plus 2pt minus 5pt%
  \abovedisplayshortskip 0pt plus 3pt%
  \belowdisplayshortskip 6pt plus 3pt minus 3pt%
  \belowdisplayskip \abovedisplayskip
  \let\@listi\@listI}
\normalsize
\renewcommand{\section}{\@startsection{section}{1}{\z@}%
  {-3.5ex \@plus -1ex \@minus -.2ex}%
  {2.3ex \@plus .2ex}%
  {\normalfont\fontsize{12pt}{14pt}\selectfont\bfseries}}
\renewcommand{\subsection}{\@startsection{subsection}{2}{\z@}%
  {-3.25ex \@plus -1ex \@minus -.2ex}%
  {1.5ex \@plus .2ex}%
  {\normalfont\fontsize{10pt}{11pt}\selectfont\bfseries}}

\title{Calibrated Weak Supervision for Post-Harvest Burned-Cropland Mapping Under Label Scarcity}

\author{%
  \begin{tabular}{c@{\qquad}c@{\qquad}c}
    \textbf{Raunak Bhagate} &
    \textbf{Maitri Polisetty} &
    \textbf{R I Minu} \\
    \small\href{mailto:rb7314@srmist.edu.in}{\textcolor{black}{\texttt{rb7314@srmist.edu.in}}} & & \\[1.4em]
    \multicolumn{3}{c}{\small SRM Institute of Science \& Technology}
  \end{tabular}%
}

\date{}

\makeatletter
\renewcommand{\@maketitle}{%
  \newpage
  \null
  \vskip 0em%
  \begin{center}%
    \let\footnote\thanks
    \rule{\linewidth}{2.6pt}\par
    \vskip 0.8em%
    {\fontsize{17pt}{20pt}\selectfont\bfseries \@title \par}%
    \vskip 0.7em%
    \rule{\linewidth}{0.5pt}\par
    \vskip 1.4em%
    {\large
      \lineskip .5em%
      \@author\par}%
    \vskip 1em%
    {\large \@date}%
  \end{center}%
  \par
  \vskip 1.5em%
}
\renewenvironment{abstract}{%
  \begin{center}%
    {\fontsize{12pt}{14pt}\selectfont\bfseries\abstractname\par}%
  \end{center}%
  \begin{quotation}%
    \fontsize{10pt}{11pt}\selectfont
}{%
  \end{quotation}%
}
\makeatother

\begin{document}

\maketitle

\begin{abstract}
Fine-resolution burned-cropland information is needed to monitor post-harvest crop-residue burning in Punjab, India, where many fires are small, fragmented, and difficult to characterize using a coarse burned-area product alone. Reliable exhaustive labels are scarce, while labels derived from active-fire detections and spectral rules can be noisy, spatially biased, and partly circular. We therefore developed a calibrated weakly supervised framework using Sentinel-2 spectral change, Visible Infrared Imaging Radiometer Suite (VIIRS) active-fire context, and MODIS MCD64A1 as a coarse external calibration and agreement reference. Three pseudo-label recipes of increasing selectivity, five feature representations, and classifier families spanning linear, tree-based, boosted, and neural models were evaluated using nested district-held-out cross-validation over three seeds and five folds. The best trained configuration combined the very-strict recipe, a multilayer perceptron, and the full optical feature set, with mean Cohen's kappa of 0.395, area under the receiver operating characteristic curve of 0.753, F1 of 0.747, and balanced accuracy of 0.703. Random Forest, XGBoost, and LightGBM were practically tied with the nominal winner. Stricter recipes achieved higher agreement with their own held-out pseudo-labels, without establishing improved label correctness or independent burned-area accuracy. Optical coverage ended before the seasonal fire context, so early burns may affect the earlier composite and late-season scars are not observed by the later composite. For deployment in Google Earth Engine, a Random Forest using the very-strict recipe and full optical feature set was retained and externally calibrated. A threshold of 0.60 produced district-level MODIS agreement of $R^2=0.636$, a mapped-to-MODIS burned-area ratio of 1.005, and Spearman correlation of 0.779 with district fire counts; pixel-level MODIS agreement remained modest ($F_1=0.205$; $\kappa=0.093$). Zero-shot transfer to Haryana was promising (mean $\kappa=0.641$), whereas Punjab cross-year stability was weak and simple Sentinel-1/Sentinel-2 feature concatenation did not improve the optical baseline. The framework is therefore most defensible for district-scale burden assessment and hotspot screening, rather than exact scar-boundary delineation or temporally invariant annual mapping. The code and data are publicly available in the \href{https://github.com/raun6k/weak-supervision-burned-cropland}{\textcolor{blue}{GitHub repository}}.
\end{abstract}

\section{Introduction}

Post-harvest crop-residue burning is a recurrent land-management practice in northwestern India and an important target for satellite-based environmental monitoring. Punjab's intensive rice--wheat system generates large quantities of seasonal residue, and burning contributes to severe episodic air pollution while complicating mitigation and policy evaluation \citep{bhuvaneshwari2019crop,jethva2019crop,lan2022airquality}. Spatially explicit burned-cropland information is consequently needed for district-scale burden assessment, hotspot identification, emissions analysis, and evaluation of residue-management interventions. Moderate-resolution burned-area products such as MODIS MCD64A1 provide consistent regional records, but their spatial support can underresolve small and fragmented agricultural burns \citep{giglio2018mcd64a1,roteta2019sentinel2}. Sentinel-2 therefore offers a useful fine-resolution representation of rapid post-fire surface change, provided that the mapping workflow can operate under realistic constraints on reference-label availability.

Agricultural burning is a difficult discrimination problem because the signal of fire is embedded within a rapidly changing post-harvest surface. Crop removal, bare-soil exposure, tillage, moisture variation, senescence, and residue burning can produce overlapping optical responses over short periods. Burns may also be discontinuous and quickly modified, reducing the temporal persistence of scars. These characteristics limit direct transfer from generic wildfire mapping and increase the need for cropland-specific context \citep{singh2009cropburning}. At the same time, exhaustive manually interpreted burn maps are expensive to create at the spatial and temporal scales required for supervised learning.

Weak supervision offers a practical alternative by constructing labels from indirect evidence such as active-fire detections, spectral rules, and pre-existing products. The central risk is that proxy labels are neither neutral nor equally informative: their thresholds, spatial buffers, and negative-sampling rules define the learning problem itself. In addition, a burn-sensitive index used to generate labels can create circularity if the same variable is reused as a predictor. Internal performance against held-out weak labels may then measure reproduction of the labeling recipe rather than a deployment-ready burned-cropland map. Spatially random train--test splits can further exaggerate generalization because nearby remote-sensing samples are dependent \citep{karasiak2022spatial,roberts2017crossvalidation}.

This study addresses these problems through a controlled weak-supervision and calibration framework for Punjab, with Haryana used only for geographic transfer testing. Sentinel-2 provides the primary optical representation; VIIRS detections provide spatial fire context for label construction; MODIS MCD64A1 provides a coarse external reference for threshold calibration and agreement diagnostics rather than pixel-perfect ground truth; and district fire counts provide an independent concordance signal at aggregate scale. Direct label-construction variables, including dNBR and NBR, are excluded from classifier inputs.

The study asks three linked questions. First, how strongly do pseudo-label selectivity, feature representation, and classifier family influence district-held-out performance under matched sampling and evaluation? Second, does the internally selected decision threshold yield a spatially plausible statewide product, or is a separate external deployment-threshold calibration required? Third, how does the selected framework behave under neighboring-state transfer, cross-year application in Punjab, and simple Sentinel-1/Sentinel-2 feature fusion? By separating weak-label discrimination from map-level calibration, the study aims to define both the achievable signal and the defensible application scale of burned-cropland mapping under label scarcity.

\section{Literature Review}

Fine-resolution burned-area mapping has increasingly moved from coarse global products toward Landsat- and Sentinel-based multitemporal change detection. MCD64A1 remains a major global reference because it combines spectral change with active-fire information and provides a consistent 500~m record \citep{giglio2018mcd64a1}. However, small-fire studies have repeatedly shown that moderate-resolution products omit or aggregate burns that can be recovered with 10--30~m imagery. Sentinel-2 algorithms use temporal change, region growing, or active-fire-seeded candidate areas to delineate smaller scars \citep{roteta2019sentinel2,sali2021burnedarea}, while Landsat/Sentinel combinations and cloud-computing workflows support broader high-resolution coverage \citep{long2019burnedarea,roy2019landsat8}. These developments are directly relevant to agricultural fires, where individual fields are often smaller than the support of coarse burned-area products.

Active-fire observations and optical burn scars provide complementary evidence rather than interchangeable labels. VIIRS offers high-frequency thermal detections at nominal 375~m resolution and is useful for locating likely fire activity in space and time \citep{schroeder2014viirs}. Optical imagery offers finer spatial structure but depends on cloud-free observations and the persistence of post-fire spectral change. Active-fire-assisted methods therefore often use hotspots as seeds or spatial priors and optical change as the delineation signal. In northwestern India, remote-sensing studies and survey-constrained analyses also show that crop-residue burning varies across districts, years, and management contexts, so neither fire counts nor a single coarse product can serve as complete pixel-level truth \citep{liu2020cropresidue,singh2009cropburning}.

Weak supervision formalizes the use of heuristic, noisy, incomplete, or correlated labeling sources when manual annotations are unavailable. Systems such as Snorkel demonstrate that labeling functions and proxy sources can create scalable training data, but their usefulness depends on source accuracy, dependence, and coverage \citep{ratner2020snorkel}. In burned-cropland mapping, threshold rules based on spectral change and proximity to active fires act as labeling functions even when not described in those terms. The resulting recipe can contain substantial predictive structure before a classifier is trained. This makes label design a scientific factor that should be varied and reported, rather than treated as a fixed preprocessing detail.

Noisy supervision also motivates comparison across model families and losses. Random Forest is widely used in remote sensing because it handles nonlinear interactions, correlated predictors, and modest sample sizes while remaining operationally simple \citep{belgiu2016randomforest,breiman2001randomforests}. XGBoost and LightGBM provide efficient boosted-tree alternatives \citep{chen2016xgboost,ke2017lightgbm}, whereas multilayer perceptrons test whether learned nonlinear combinations improve performance. Symmetric cross-entropy was proposed to reduce neural-network sensitivity to noisy labels \citep{wang2019symmetric}, but gains under remotely sensed proxy supervision remain empirical rather than guaranteed.

Evaluation design is as important as classifier choice. Spatial dependence between training and test samples can inflate accuracy, making geographically disjoint validation more realistic for regional mapping \citep{karasiak2022spatial,roberts2017crossvalidation}. A second issue is thresholding: the probability cutoff that maximizes agreement with weak labels need not yield plausible mapped area or district patterns. External evidence is therefore needed after model fitting, but the evidential role of each source must remain explicit. Coarse MODIS agreement can support regional plausibility and area calibration, yet it cannot establish exact 20~m scar boundaries. The unresolved gap is a single controlled study that jointly tests pseudo-label design, circularity-aware features, classifier families, spatially held-out evaluation, external deployment-threshold calibration, and transfer behavior for fine-resolution burned cropland in Punjab.

\section{Materials and Methods}

\subsection{Study Area, Temporal Scope, and Data Roles}

\begin{figure}[H]
  \centering
  \includegraphics[height=1.9in,keepaspectratio]{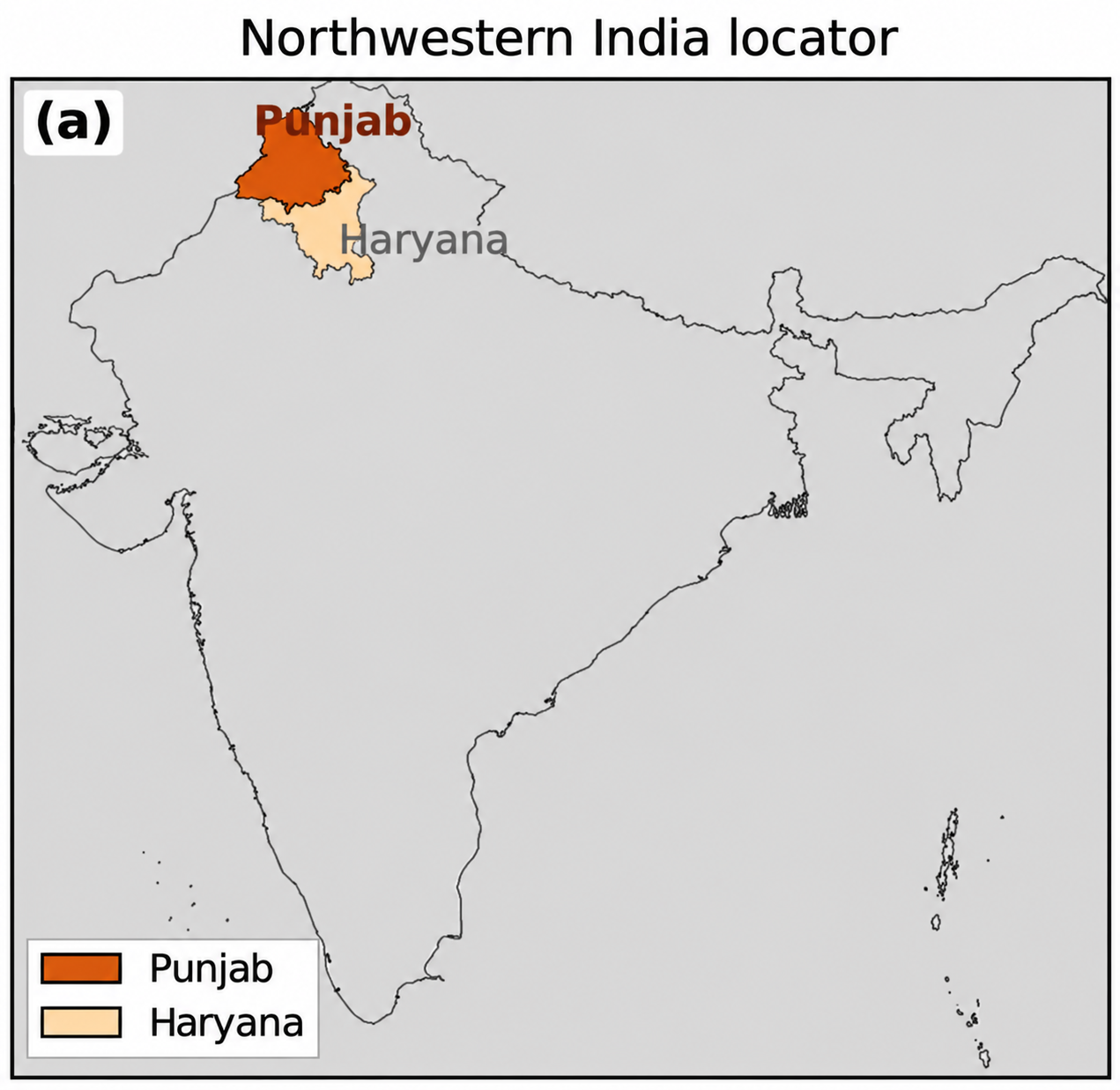}\hfill
  \includegraphics[height=1.9in,keepaspectratio]{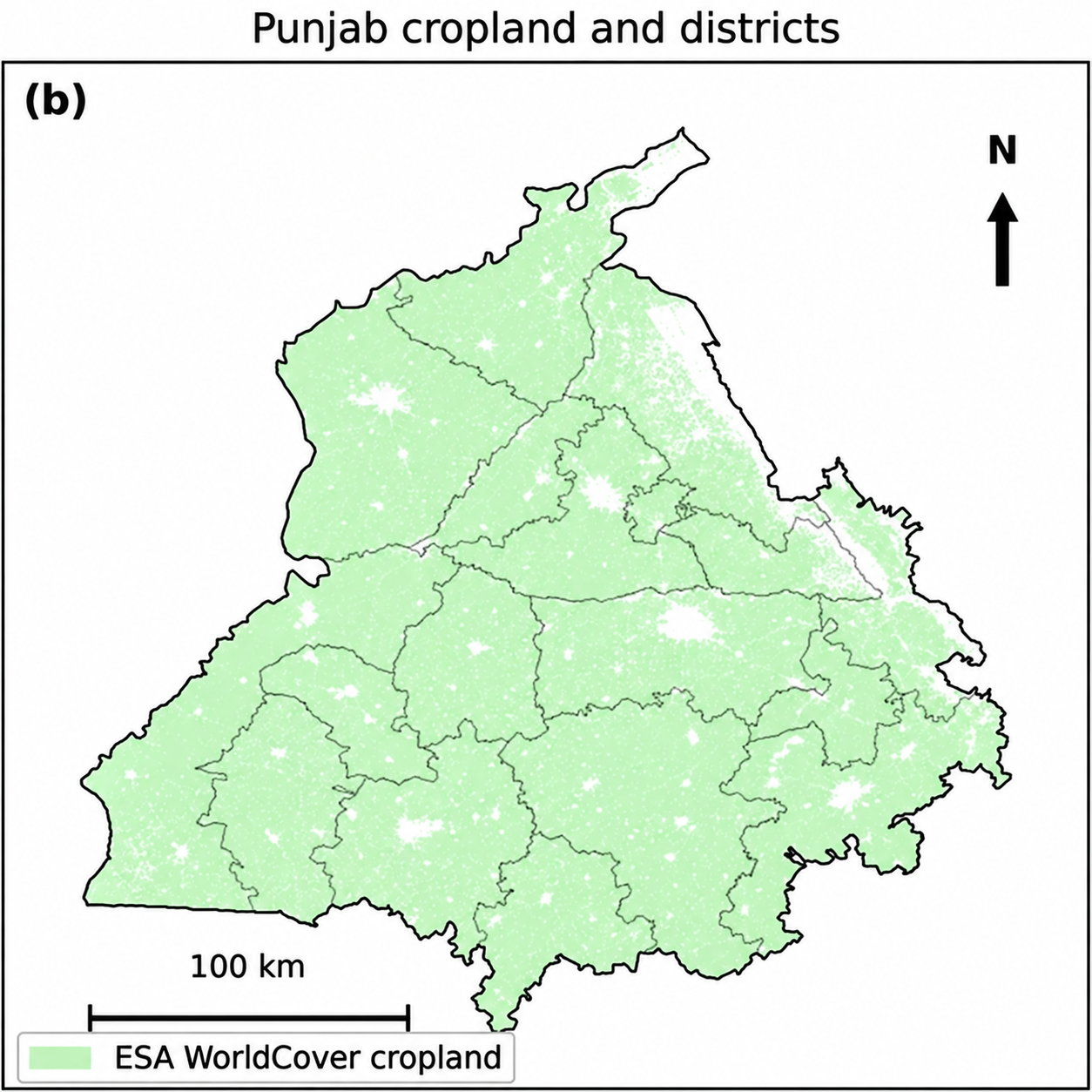}\hfill
  \includegraphics[height=1.9in,keepaspectratio]{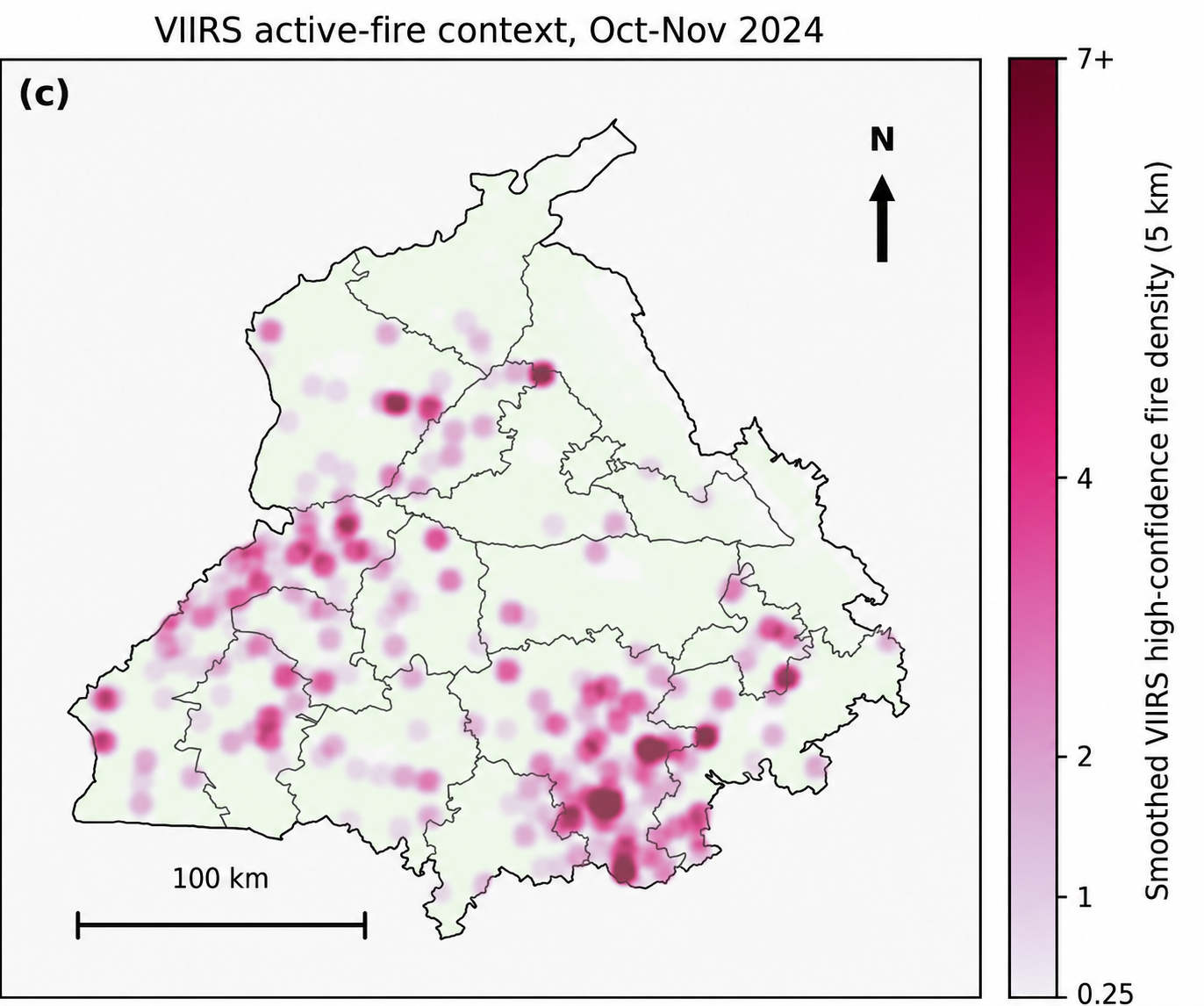}
  \caption{Study area and active-fire context. (a) Punjab and Haryana in northwestern India. (b) Punjab districts and the ESA WorldCover cropland mask used to restrict analysis. (c) High-confidence VIIRS active-fire density over Punjab cropland from 1 October--29 November 2024.}
  \label{fig:study-area}
\end{figure}

Punjab, India, was the primary study area, and Haryana was used as a neighboring transfer region (Figure~\ref{fig:study-area}). Administrative units were obtained from FAO GAUL level-2 boundaries. All label construction, sampling, and inference were restricted to cropland using ESA WorldCover 2021 v200 class 40 \citep{zanaga2022worldcover}. This restriction kept the task specific to burned cropland rather than generic fire detection and retained district identity for spatial folds and zonal summaries.

The primary experiment used the 2024 post-monsoon residue-burning season. The earlier optical composite covered 20 September--9 October 2024, and the later optical composite covered 15 October--4 November 2024, accounting for Earth Engine's exclusive end dates. VIIRS fire context covered 1 October--29 November, while MODIS used the October and November monthly products. The optical observation windows were therefore shorter than the seasonal reference period. Punjab 2023 was processed with the same temporal structure for the reported cross-year comparison. Sentinel-2 surface reflectance was the primary optical input \citep{drusch2012sentinel2}; Sentinel-1 ground-range-detected imagery was used only in the fusion experiment \citep{torres2012sentinel1}; SNPP and NOAA-20 VIIRS detections provided active-fire context; MODIS MCD64A1 provided the external calibration and agreement reference; and district fire counts provided a rank-concordance signal.

\FloatBarrier

\subsection{Framework Overview and Image Preprocessing}

\begin{figure}[H]
  \centering
  \includegraphics[width=\linewidth]{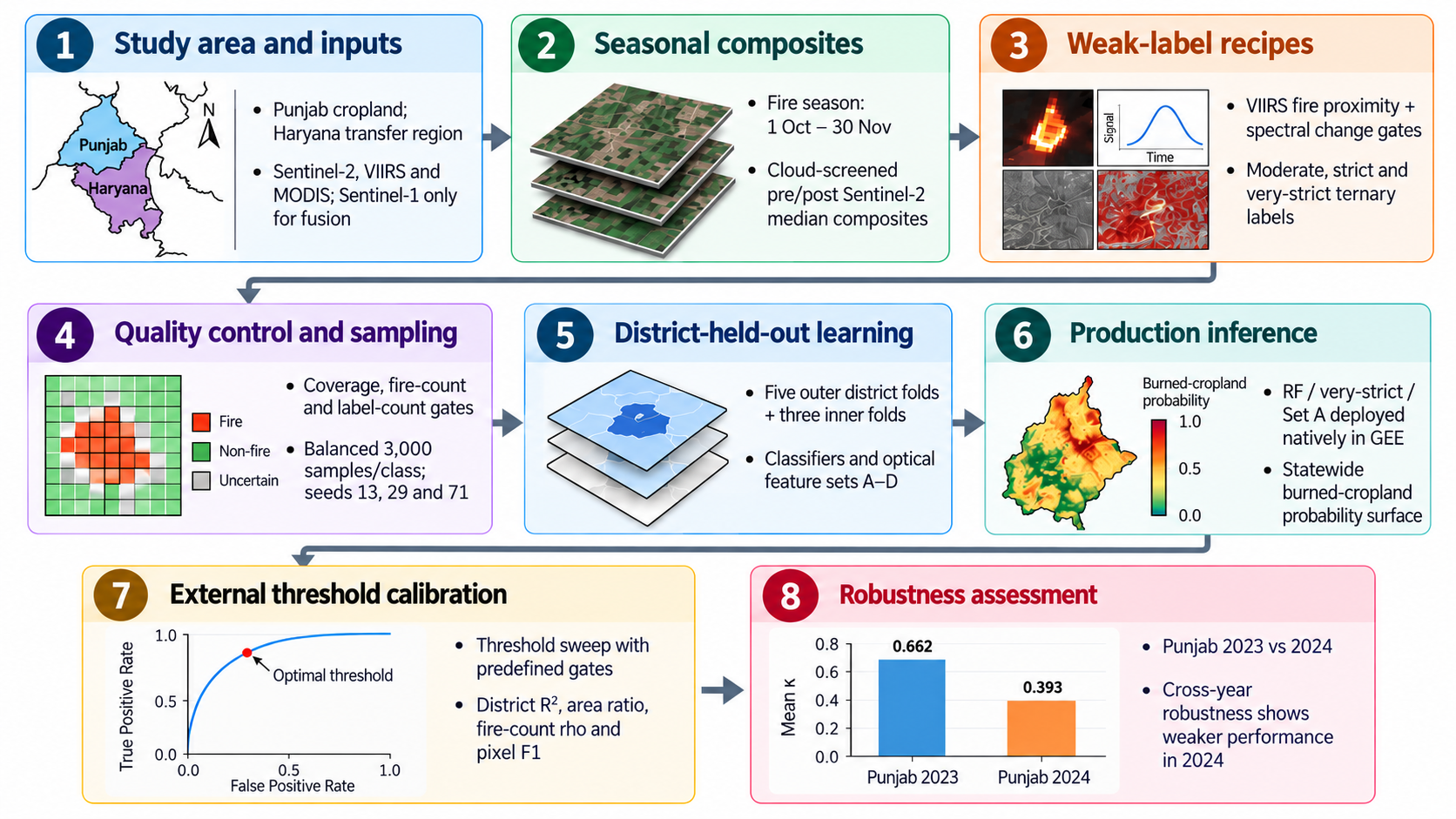}
  \caption{End-to-end weak-supervision and calibration workflow. Sentinel-2 preprocessing, VIIRS-anchored pseudo-labels, matched district-held-out evaluation, statewide RF inference, external deployment-threshold calibration, and robustness testing are kept as distinct stages. The schematic's pre/post labels denote the earlier/later optical composites; its October--November season denotes the broader reference period. NBR and dNBR are used for label construction but excluded from classifier inputs.}
  \label{fig:workflow}
\end{figure}

The workflow consisted of seasonal compositing, weak-label generation, quality-control gating, matched sampling, district-held-out model evaluation, statewide probability inference, external threshold calibration, and robustness testing (Figure~\ref{fig:workflow}). Google Earth Engine was used for geospatial assembly, preprocessing, weak-label construction, and map export; Python was used for classifier benchmarking, paired statistics, calibration analysis, and reporting.

Sentinel-2 acquisitions were filtered by date and study area, and scenes with \texttt{CLOUDY\_PIXEL\_}\allowbreak\texttt{PERCENTAGE} $\leq 80$ were retained. A Sentinel-2 cloud-probability threshold of 35 was applied together with QA60 and Scene Classification Layer screening for cloud, cirrus, shadow, and invalid observations. Median composites were generated for the earlier and later optical windows. Ten bands were retained (B2, B3, B4, B5, B6, B7, B8, B8A, B11, and B12), together with normalized difference vegetation index (NDVI), normalized difference moisture index (NDMI), normalized difference red-edge index (NDRE), and temporal differences. NBR and dNBR were calculated for label construction but excluded from every classifier feature set. The raw SWIR2 change term was $dB12=B12_{\mathrm{later}}-B12_{\mathrm{earlier}}$. Sentinel-1 preprocessing used GRD scenes in IW mode, descending orbit, and VV/VH polarization; median earlier and later composites and later-minus-earlier delta backscatter were exported at 20~m for the dedicated fusion test. Before model training, plausibility gates required at least 95\% cropland coverage in each optical composite, at least 1,000 high-confidence VIIRS detections, a nonzero MODIS burned reference, and at least 10,000 statewide positive weak labels per recipe. Weak positives were also required to reach an overlap score of at least 0.25 with MODIS at 500~m support, while district label availability and sampled feature-correlation checks were screened before training.

\FloatBarrier

\subsection{Weak-Label Recipes and Feature Design}

\begin{figure}[H]
  \centering
  \begin{minipage}[t]{0.31\linewidth}
    \vspace{0pt}\centering
    \includegraphics[width=\linewidth,height=1.50in,keepaspectratio]{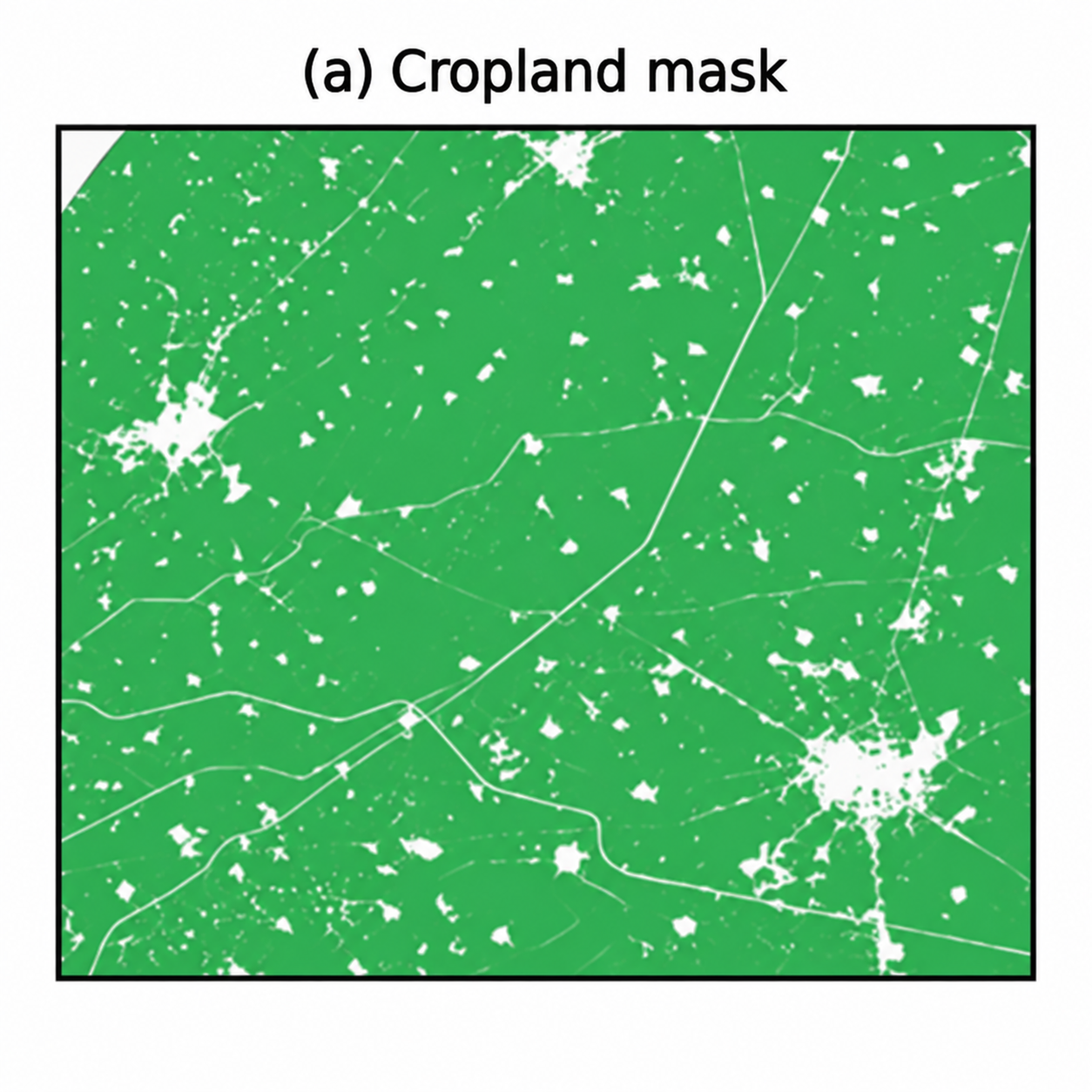}
  \end{minipage}\hfill
  \begin{minipage}[t]{0.31\linewidth}
    \vspace{0pt}\centering
    \includegraphics[width=\linewidth,height=1.50in,keepaspectratio]{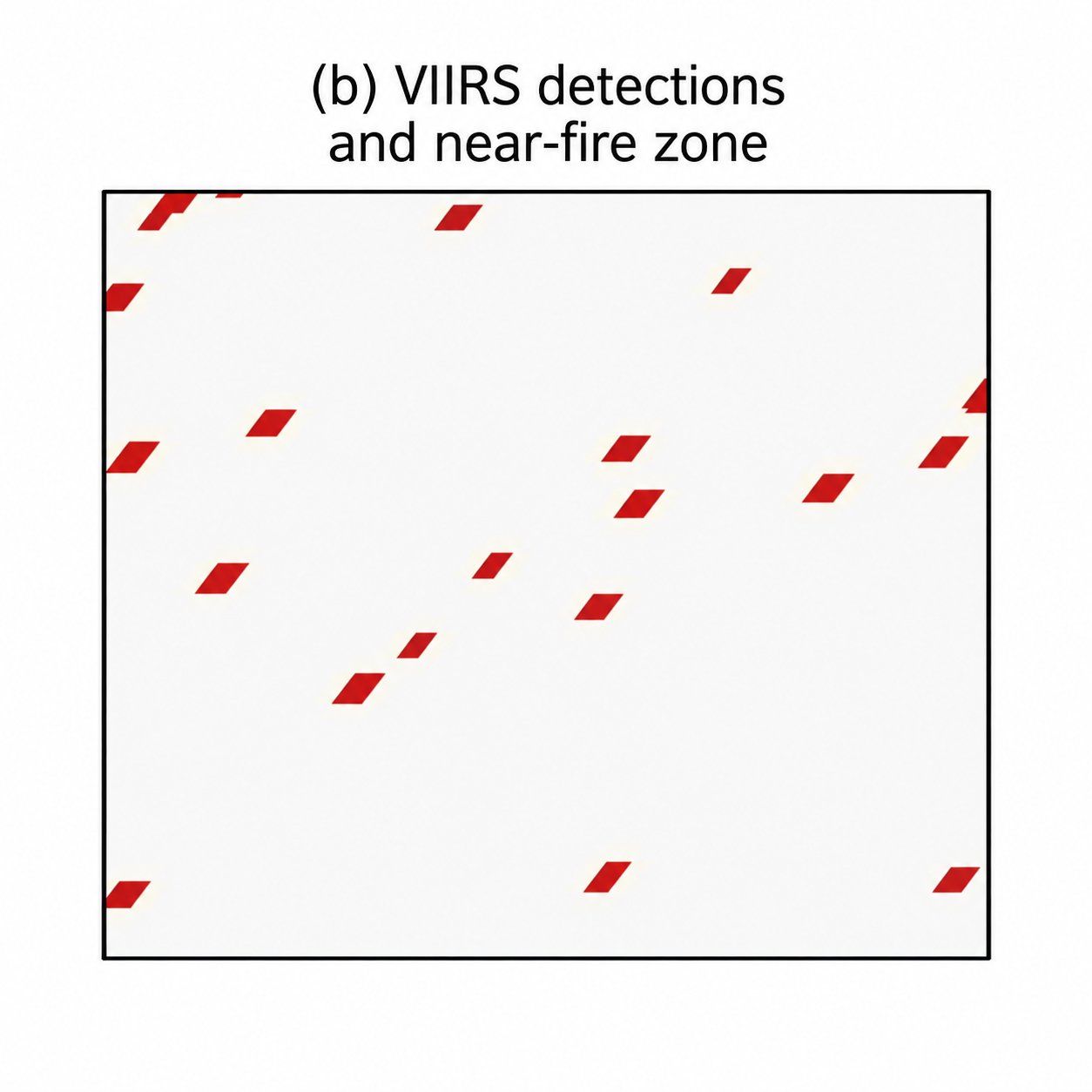}
  \end{minipage}\hfill
  \begin{minipage}[t]{0.31\linewidth}
    \vspace{0pt}\centering
    \includegraphics[width=\linewidth,height=1.50in,keepaspectratio]{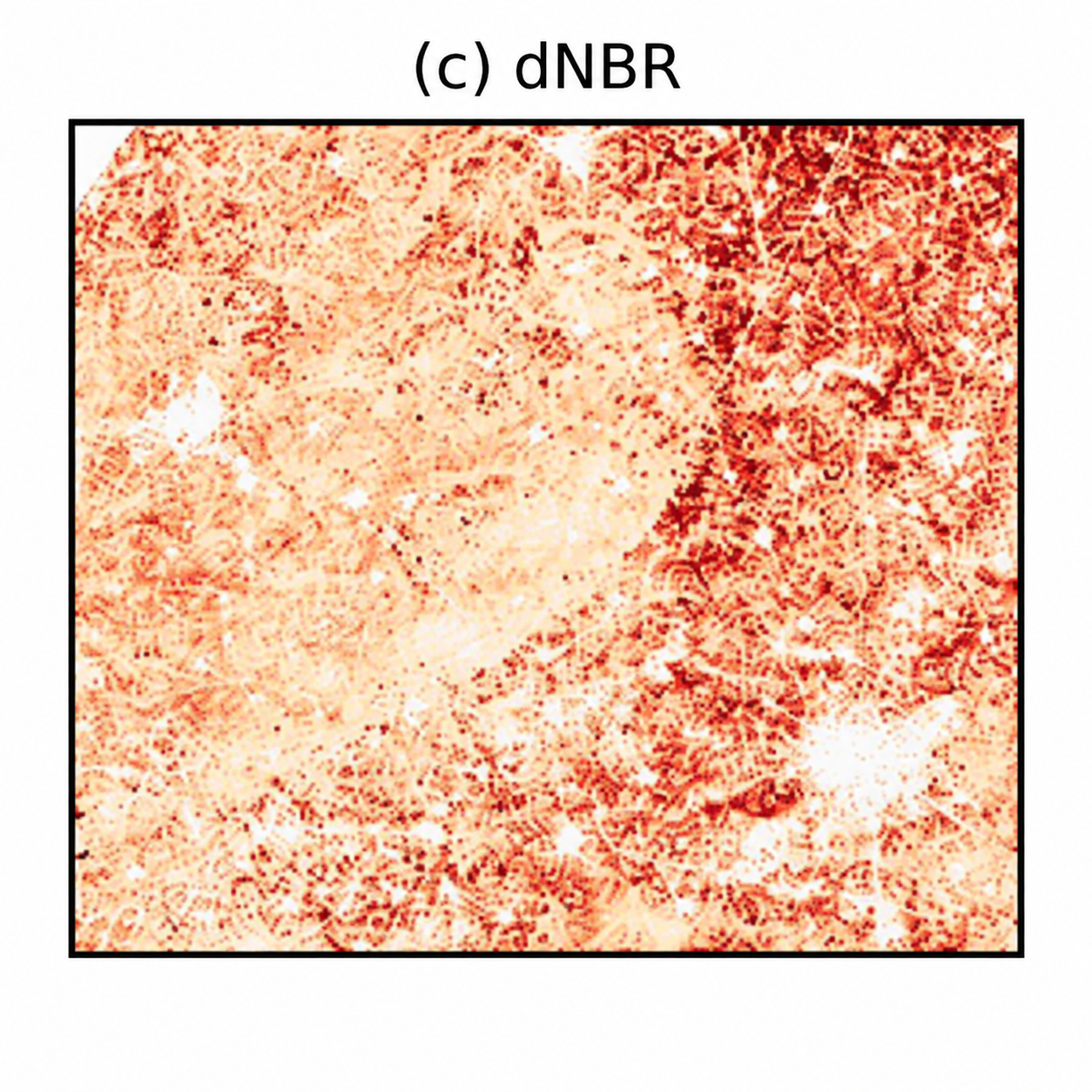}
  \end{minipage}

  \vspace{0.25em}

  \begin{minipage}[t]{0.31\linewidth}
    \vspace{0pt}\centering
    \includegraphics[width=\linewidth,height=1.50in,keepaspectratio]{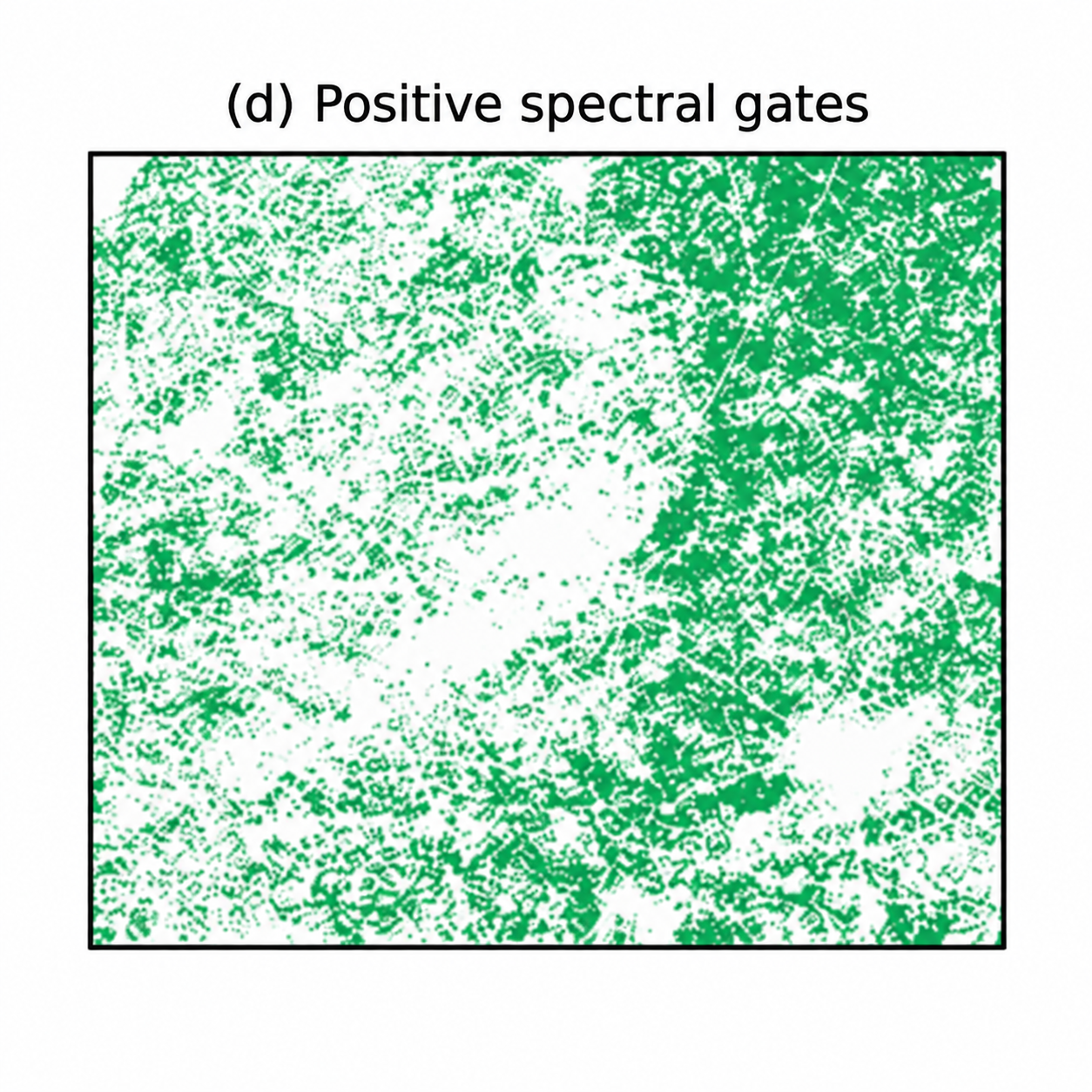}
  \end{minipage}\hfill
  \begin{minipage}[t]{0.31\linewidth}
    \vspace{0pt}\centering
    \includegraphics[width=\linewidth,height=1.50in,keepaspectratio]{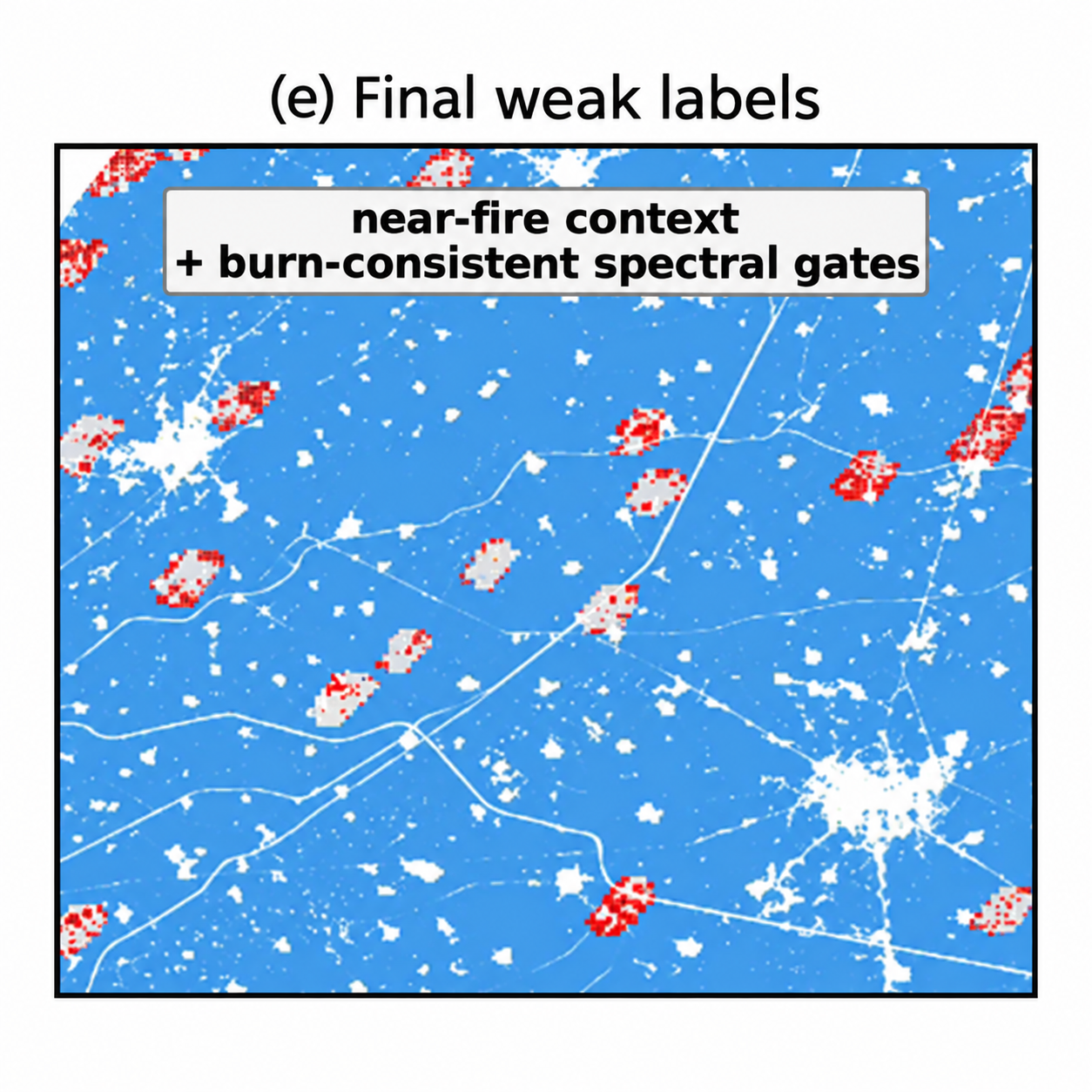}
  \end{minipage}\hfill
  \begin{minipage}[t]{0.31\linewidth}
    \vspace{0pt}\centering
    \includegraphics[width=\linewidth,height=1.50in,keepaspectratio]{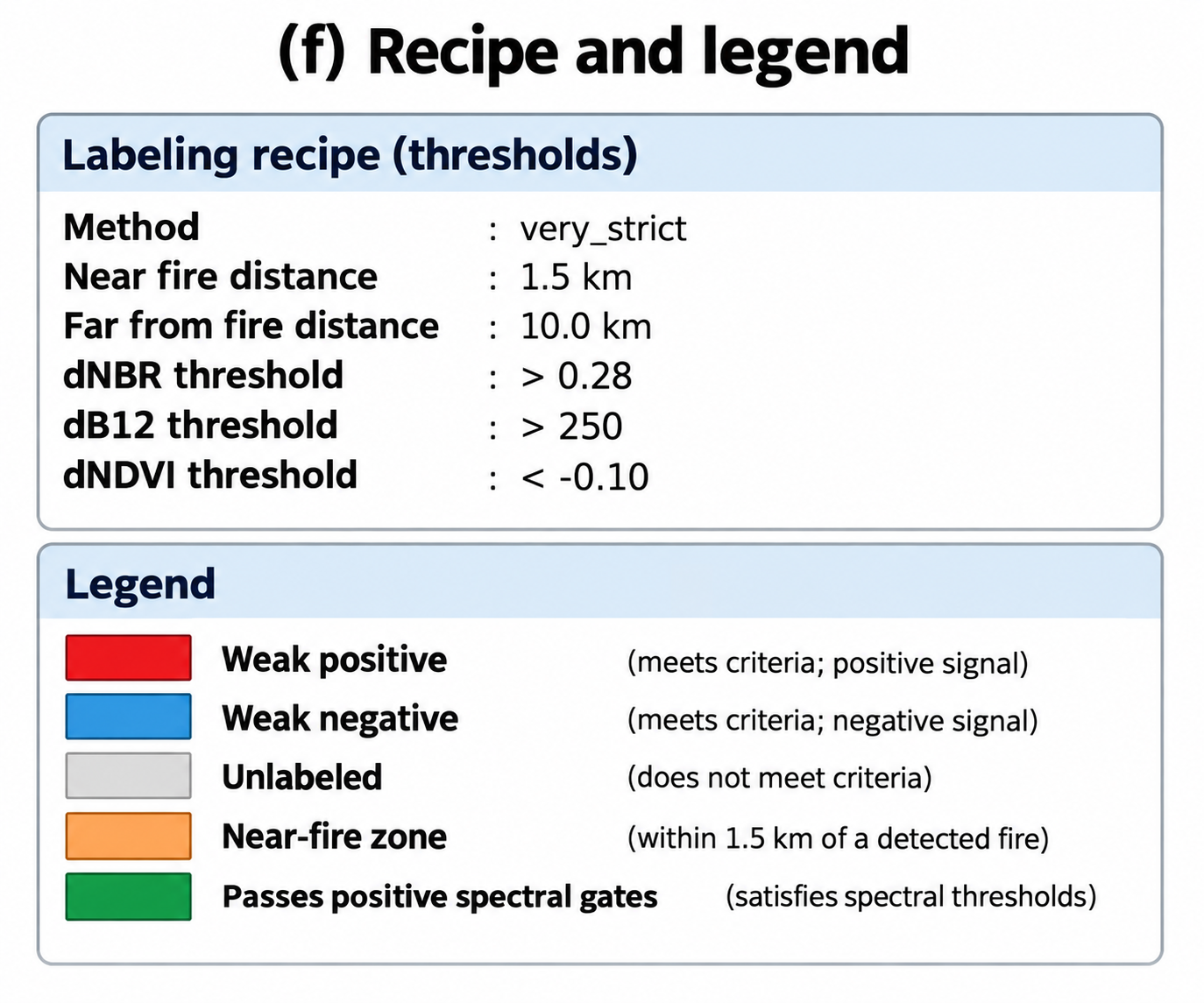}
  \end{minipage}
  \caption{Weak-label construction. Cropland and VIIRS fire context are intersected with burn-consistent spectral gates to produce weak positives, weak negatives, and unlabeled pixels. The displayed parameter block corresponds to the exact very-strict configuration.}
  \label{fig:weak-labels}
\end{figure}

Weak labels combined VIIRS fire proximity with burn-consistent optical change. High-confidence SNPP and NOAA-20 detections were merged for the VIIRS context window, converted to a fire-context layer, and transformed into distance to the nearest detection. Figure~\ref{fig:weak-labels} illustrates the resulting ternary construction.

Each cropland pixel was assigned as weak positive, weak negative, or unlabeled. Positive pixels had to fall within a recipe-specific near-fire distance and pass dNBR and SWIR change gates; the most selective recipe also required an NDVI decrease. Negative pixels were either sufficiently far from detected fire or near fire but spectrally quiet. Pixels satisfying neither class rule were excluded from training.

Three recipes represented an explicit quantity--selectivity axis (Table~\ref{tab:recipes}). The moderate recipe used the widest near-fire buffer and lowest dNBR threshold. The strict recipe tightened the spatial and spectral conditions. The very-strict recipe reduced the near-fire radius to 1.5~km, increased the dNBR and $dB12$ requirements, extended the far-from-fire distance, tightened hard-negative limits, and added an NDVI-decrease gate. The recipes shared the same source layers and seasonal windows; only thresholds, buffers, and gates changed.

Five feature sets were defined by analytical purpose (Table~\ref{tab:features}). Set A was the full 39-variable optical representation with earlier, later, and delta information. Set B retained 13 delta-only variables. Set C combined earlier optical state and deltas. Set D removed SWIR-related predictors as the primary circularity ablation. Set E augmented Set A with six Sentinel-1 variables. Set A served as the exported superset for optical comparisons, so Sets B--D were reconstructed by column selection on identical samples.

\begin{table}[H]
  \centering
  \normalsize
  \renewcommand{\arraystretch}{1.08}
  \setlength{\tabcolsep}{3pt}
  \begin{tabular}{@{}>{\raggedright\arraybackslash}p{0.19\linewidth} >{\centering\arraybackslash}p{0.08\linewidth} >{\centering\arraybackslash}p{0.10\linewidth} >{\centering\arraybackslash}p{0.09\linewidth} >{\centering\arraybackslash}p{0.09\linewidth} >{\centering\arraybackslash}p{0.09\linewidth} >{\centering\arraybackslash}p{0.08\linewidth} >{\centering\arraybackslash}p{0.10\linewidth}@{}}
    \toprule
    Recipe & \shortstack{d\\NBR} & \shortstack{Near\\km} & \shortstack{$dB$\\12} & \shortstack{HN\\d} & \shortstack{HN\\b} & \shortstack{Far\\km} & NDVI \\
    \midrule
    Moderate & 0.18 & 3.0 & $>200$ & 0.06 & 120 & 8 & None \\
    Strict & 0.22 & 2.0 & $>200$ & 0.05 & 100 & 8 & None \\
    Very-strict & 0.28 & 1.5 & $>250$ & 0.04 & 80 & 10 & $>0.10$ \\
    \bottomrule
  \end{tabular}
  \caption{Weak-label recipe parameters. Near- and far-fire distances are in km; HN denotes hard-negative, with $d$ and $b$ denoting dNBR and $dB12$. The recipes share the same inputs and seasonal windows and differ only in spatial and spectral selection constraints.}
  \label{tab:recipes}
\end{table}

\begin{table}[H]
  \centering
  \normalsize
  \renewcommand{\arraystretch}{1.08}
  \setlength{\tabcolsep}{2pt}
  \begin{tabular}{@{}>{\centering\arraybackslash}p{0.08\linewidth} >{\centering\arraybackslash}p{0.09\linewidth} >{\raggedright\arraybackslash}p{0.36\linewidth} >{\raggedright\arraybackslash}p{0.39\linewidth}@{}}
    \toprule
    Set & Size & Composition & Analytical purpose \\
    \midrule
    A & 39 & Earlier, later, and delta values for 10 Sentinel-2 bands plus three indices & Full optical production representation \\
    B & 13 & Delta-only optical variables & Tests whether change alone is sufficient \\
    C & 26 & Earlier optical state plus delta variables & Retains baseline context with reduced redundancy \\
    D & 33 & Full optical set without SWIR-related predictors & Primary circularity ablation \\
    E & 45 & Set A plus Sentinel-1 VV/VH earlier, later, and delta & Simple optical--radar fusion test \\
    \bottomrule
  \end{tabular}
  \caption{Feature-set definitions. NBR and dNBR are excluded from every classifier feature set because they are used in weak-label construction.}
  \label{tab:features}
\end{table}

\FloatBarrier

\subsection{Sampling, Classifiers, Baselines, and District-Held-Out Evaluation}

For each recipe and seed, stratified sampling selected 3,000 weak-positive and 3,000 weak-negative pixels. Three fixed seeds (13, 29, and 71) were used. Districts, rather than pixels, were assigned to five outer folds so that all samples from a district were held out together. The resulting 15 seed--fold evaluations per configuration reduced local spatial leakage and measured transfer to unseen districts. Within each outer training set, three-fold district-based inner validation selected a probability threshold from 0.1 to 0.9 in steps of 0.1 by maximizing Cohen's kappa. Held-out districts were not used for threshold tuning.

The learned-model comparison included Logistic Regression, Random Forest (RF), Gradient Boosted Trees, XGBoost, LightGBM, a multilayer perceptron (MLP), and an MLP trained with symmetric cross-entropy (MLP-SCE). RF used 300 trees, square-root feature subsampling, and minimum leaf size 1. Gradient Boosted Trees used 300 estimators, a learning rate of 0.1, and at most 64 leaf nodes. The baseline MLP used hidden layers of 128 and 64 units, batch normalization, ReLU activation, dropout of 0.3, Adam optimization, learning rate 0.001, batch size 256, and 200 epochs. MLP-SCE used the same architecture with symmetric cross-entropy weights $\alpha=1$ and $\beta=1$. All classifier families were evaluated on the same matched samples and district folds. Valid non-learned benchmark comparators were the recipe-rule threshold and a fixed dNBR threshold of 0.12. MODIS was not treated as a classifier baseline because a valid district-held-out MODIS benchmark was not completed.

Internal evaluation was explicitly against held-out weak pseudo-labels, not human-scored pixel truth. Each recipe defined its own evaluation labels and sampled population, so cross-recipe differences do not isolate training-label quality against a common independent reference. Cohen's kappa was the primary ranking metric, supported by $F_1$, balanced accuracy, area under the receiver operating characteristic curve (AUC), precision, recall, specificity, selected threshold, and predicted positive rate. Decision-relevant paired contrasts used Wilcoxon signed-rank tests across matched seed--fold splits, Bonferroni correction, and Cliff's delta.

\subsection{Production Model and External Deployment-Threshold Calibration}

The best trained configuration and the production configuration were separated. The nominal best learned model was MLP/very-strict/Set A. RF/very-strict/Set A was retained for production because the mapping workflow constructed the model directly in Google Earth Engine, whose exporter supports native RF and gradient-boosted trees but not the PyTorch MLP. The MLP advantage over RF on Set A was only 0.00253 kappa and was non-significant after correction (corrected $p=1.0$; Cliff's $\delta=-0.013$). RF therefore avoided a separate neural inference path without a meaningful performance penalty.

Feature Set B was marginally stronger than Set A for RF in internal cross-validation and was more parsimonious. Set A was nevertheless the actual full-feature production configuration because it had been manually fixed, exported, and taken through the complete statewide calibration workflow. Set A is therefore reported as the deployed configuration, not as an internally optimal RF feature set.

After retraining RF/very-strict/Set A on all labeled Punjab samples, the statewide probability surface was evaluated at 17 external candidate thresholds from 0.10 to 0.90 in 0.05 increments. Three deployment gates were defined: district-level $R^2$ against MODIS of at least 0.50, mapped-to-MODIS burned-area ratio of at most 2.0, and district fire-count Spearman correlation of at least 0.60. Thresholds were ranked by: (1) maximum gates passed; (2) preference for passing the area-ratio gate; (3) area ratio closest to 1 on a log scale; (4) higher district $R^2$; (5) higher fire-count Spearman correlation; (6) higher pixel-level MODIS $F_1$; and (7) lower threshold. For pixel-level agreement diagnostics, the 20~m model output was brought to the 500~m MODIS grid before overlap metrics were computed, avoiding direct pixel matching across unequal supports. Thresholds 0.55, 0.60, 0.65, and 0.70 passed all three gates; 0.60 was selected because its area ratio, 1.005, was closest to parity. MODIS and fire-count values at the selected threshold are calibration statistics, and pixel metrics are agreement diagnostics on the calibration reference rather than independent post-selection validation.

\subsection{Robustness Experiments}

Geographic transfer used three RF/very-strict/Set A models fitted on Punjab samples, one per seed, and applied directly to Haryana. No Haryana training, fine-tuning, or threshold calibration was performed. Each model used its Punjab-only inner-cross-validation threshold: 0.30, 0.40, and 0.30. The reported Haryana kappa is the mean sample-level zero-shot performance across the three models.

Temporal stability compared Punjab 2023 with Punjab 2024 under the same seasonal design and evaluation framework. The paper reports only these two years. Sensor augmentation compared optical Set A with Set E, which added earlier, later, and delta VV/VH features through simple concatenation. The result therefore tests this specific fusion design, not the general utility of radar for burned-area mapping.

\section{Results}

\subsection{Recipe Selectivity and Held-Out Pseudo-Label Agreement}

\begin{figure}[H]
  \centering
  \includegraphics[width=0.96\linewidth]{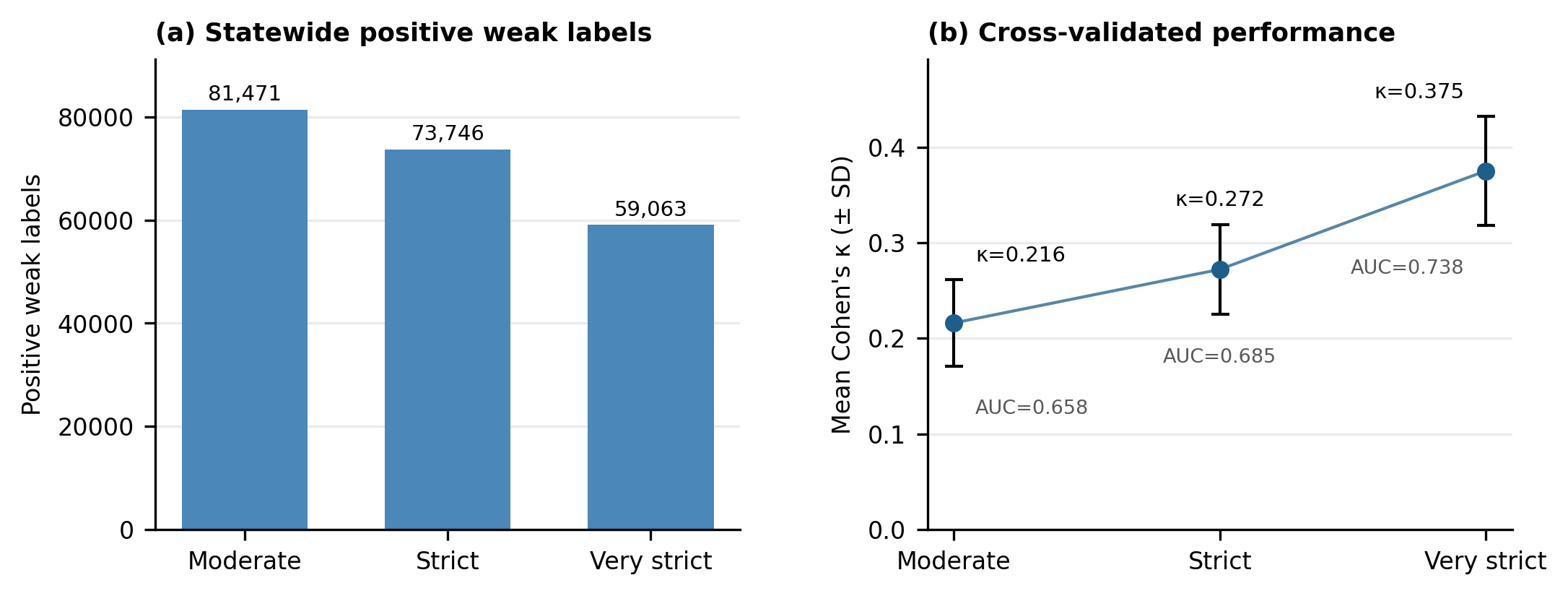}
  \caption{Quantity--selectivity pattern across pseudo-label recipes. The very-strict recipe produced fewer positive weak labels and higher mean district-held-out agreement with its own pseudo-labels. Evaluation labels and sampled populations varied across recipes, so this pattern does not establish improved label correctness or independent burned-area accuracy.}
  \label{fig:recipe-selectivity}
\end{figure}

Stricter recipes achieved higher agreement with their own held-out pseudo-labels. The moderate recipe generated 81,471 positive labels and achieved mean kappa of 0.216 and mean AUC of 0.658 across classifiers and feature sets. The strict recipe generated 73,746 positives and improved to kappa 0.272 and AUC 0.685. The very-strict recipe generated the fewest positives (59,063) but reached mean kappa 0.375 and AUC 0.738 (Figure~\ref{fig:recipe-selectivity}). These increases describe agreement on recipe-specific evaluation populations, rather than an independently verified improvement in burned-area accuracy.

The best trained configuration was MLP/very-strict/Set A, with mean kappa 0.3951, AUC 0.753, $F_1$ 0.747, and balanced accuracy 0.703 across 15 district-held-out splits (Table~\ref{tab:leading-configs}). Several alternatives were effectively co-ranked: RF/very-strict/Set B reached kappa 0.395, XGBoost/very-strict/Set A reached 0.394, RF/very-strict/Set A reached 0.3926, and leading LightGBM variants were within the same narrow range. The paired MLP/A versus RF/A comparison yielded $\Delta\kappa=0.00253$, corrected $p=1.0$, and Cliff's $\delta=-0.013$, providing no evidence of a meaningful advantage for the neural model.

The non-learned comparisons further showed how much signal was already embedded in label construction. The fixed dNBR baseline remained weak at approximately kappa 0.126--0.127 across recipes. In contrast, the recipe-rule comparator improved with recipe strength and reached kappa 0.426 for the very-strict recipe. This comparator is structurally close to the pseudo-label construction and is not a learned model, and its performance reflects agreement with recipe-specific pseudo-labels rather than independently measured label quality. Learning provided a transferable probability surface and multivariate refinement, while recipe comparisons remained conditional on their respective evaluation labels.

Circularity-aware feature analysis supported a broader optical-change signal rather than direct reproduction of dNBR. NBR and dNBR were absent from all model inputs, and performance declined for the no-SWIR Set D, showing that burn-sensitive optical information beyond the excluded label-construction indices contributed to discrimination. Set B's competitive RF result also showed that temporal deltas carried substantial predictive information, although the production workflow retained full Set A for operational continuity. Together, these ablations support a multivariate optical-change interpretation while avoiding a stronger claim about individual-variable importance.

\begin{table}[H]
  \centering
  \normalsize
  \renewcommand{\arraystretch}{1.08}
  \setlength{\tabcolsep}{3.5pt}
  \begin{tabular}{ccccccccc}
    \toprule
    Rank & Recipe & Model & Set & Mean $\kappa$ & SD $\kappa$ & AUC & $F_1$ & Bal. acc. \\
    \midrule
    1 & Very-strict & MLP & A & 0.395 & 0.049 & 0.753 & 0.747 & 0.703 \\
    2 & Very-strict & RF & B & 0.395 & 0.053 & 0.747 & 0.744 & 0.703 \\
    3 & Very-strict & XGBoost & A & 0.394 & 0.053 & 0.740 & 0.747 & 0.703 \\
    4 & Very-strict & RF & A & 0.393 & 0.056 & 0.744 & 0.746 & 0.701 \\
    5 & Very-strict & LightGBM & B & 0.393 & 0.052 & 0.740 & 0.748 & 0.701 \\
    \bottomrule
  \end{tabular}
  \caption{Leading learned configurations. Values are means across 15 district-held-out evaluations (three seeds $\times$ five folds).}
  \label{tab:leading-configs}
\end{table}

\FloatBarrier

\subsection{External Threshold Calibration and Scale-Dependent Agreement}

\begin{figure}[H]
  \centering
  \includegraphics[width=0.82\linewidth]{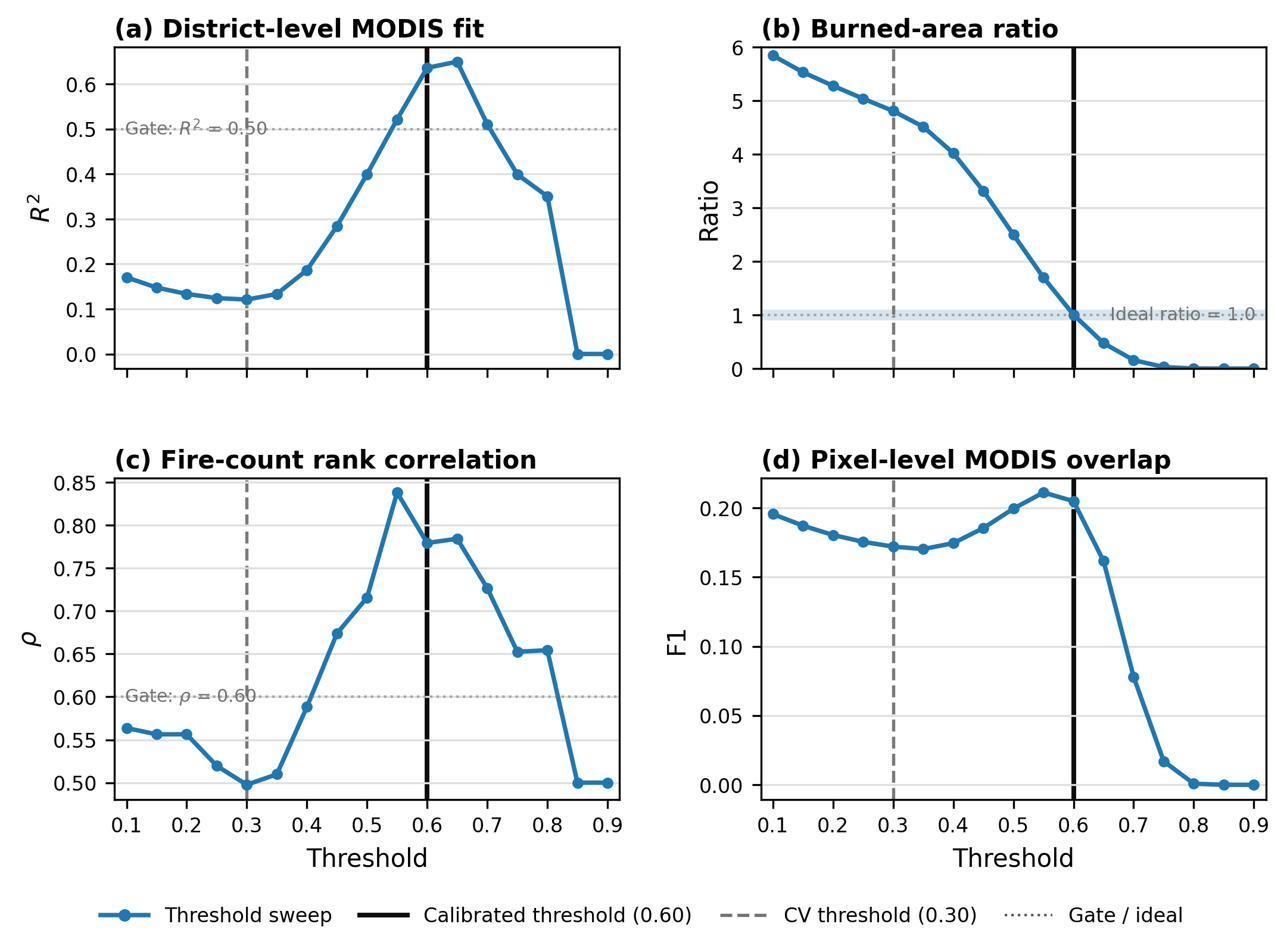}
  \caption{External deployment-threshold calibration for RF/very-strict/Set A. The internal threshold (0.30) and calibrated threshold (0.60) are compared across district MODIS fit, aggregate area ratio, fire-count rank correlation, and pixel MODIS $F_1$. The selected cutoff balances the predefined criteria rather than maximizing a single metric.}
  \label{fig:threshold-calibration}
\end{figure}

The internally selected threshold was not suitable for statewide deployment. RF/very-strict/Set A had a median cross-validation threshold of 0.30, but applying that cutoff to the full probability map produced a mapped burned-area ratio far above MODIS. External calibration shifted the decision threshold to 0.60 (Figure~\ref{fig:threshold-calibration}). This change demonstrates that discrimination among weakly labeled samples and calibration of mapped extent are distinct stages.

Thresholds 0.55--0.70 satisfied all three deployment gates. The ranking rule selected 0.60 because its mapped-to-MODIS area ratio was 1.005, closest to aggregate parity. At this threshold, district-level $R^2$ against MODIS was 0.636 and Spearman correlation with district fire counts was 0.779 (Table~\ref{tab:production-calibration}). These statistics support district-scale burden and rank consistency after calibration.

Pixel-level agreement with the coarse MODIS reference remained modest: $F_1$ was 0.205 and kappa was 0.093. Precision and recall were both approximately 0.20, while specificity was substantially higher. The result is consistent with support mismatch between a fine-resolution product and a 500~m reference, as well as with residual errors in both sources. It does not support exact scar-boundary claims.

District residuals were geographically structured. Firozpur and Moga showed close agreement, whereas Sangrur, Muktsar, Bathinda, and Mansa were underestimated relative to MODIS. Several lower-MODIS-burn districts were overestimated. The map can therefore reproduce aggregate statewide extent and much of the district ranking while retaining district-specific bias. The reported MODIS and fire-count values are the evidence used to choose the threshold and should be interpreted as calibration and agreement statistics, not as an independent evaluation conducted after threshold selection.

\begin{table}[H]
  \centering
  \normalsize
  \renewcommand{\arraystretch}{1.08}
  \setlength{\tabcolsep}{5pt}
  \begin{tabularx}{\linewidth}{@{}>{\raggedright\arraybackslash}p{0.52\linewidth} >{\centering\arraybackslash}X@{}}
    \toprule
    Item & Value \\
    \midrule
    Production classifier & Random Forest (300 trees) \\
    Pseudo-label recipe & Very-strict \\
    Feature set & A (39 optical variables) \\
    Median internal CV threshold & 0.30 \\
    Externally calibrated threshold & 0.60 \\
    District $R^2$ vs.\ MODIS & 0.636 \\
    Mapped/MODIS burned-area ratio & 1.005 \\
    Spearman $\rho$ vs.\ district fire counts & 0.779 \\
    Pixel $F_1$ vs.\ MODIS & 0.205 \\
    Pixel $\kappa$ vs.\ MODIS & 0.093 \\
    Deployment gates passed & 3/3 \\
    \bottomrule
  \end{tabularx}
  \caption{Production configuration and selected-threshold calibration/agreement statistics.}
  \label{tab:production-calibration}
\end{table}

\FloatBarrier

\subsection{Geographic Transfer, Cross-Year Stability, and Sentinel-1 Fusion}

Zero-shot transfer to Haryana was promising. The three Punjab-trained RF models, applied without Haryana training or recalibration and using Punjab-only thresholds of 0.30, 0.40, and 0.30, achieved mean sample-level kappa of 0.641. This result indicates that the learned decision surface transferred to a neighboring agricultural region under the tested seasonal and sensor conditions. It does not establish broad geographic generalization beyond similar settings.

Temporal behavior was substantially weaker. Punjab 2023 achieved kappa 0.662, whereas Punjab 2024 achieved 0.393, with a two-year standard deviation of 0.135. The framework should therefore not be described as temporally invariant. Year-specific changes in fire intensity, crop timing, atmospheric contamination, compositing quality, or pseudo-label fidelity may all contribute, but the completed experiments do not isolate a single cause.

Simple optical--radar concatenation did not improve the selected optical baseline. RF/very-strict/Set A achieved mean kappa 0.393, while Set E reached 0.378, a difference of $-0.015$ (Table~\ref{tab:robustness}). This negative result is bounded to the tested pre/post/delta VV/VH concatenation and does not imply that Sentinel-1 is generally uninformative.

\begin{table}[H]
  \centering
  \normalsize
  \renewcommand{\arraystretch}{1.08}
  \setlength{\tabcolsep}{3pt}
  \begin{tabularx}{\linewidth}{@{}>{\raggedright\arraybackslash}p{0.19\linewidth} >{\raggedright\arraybackslash}p{0.38\linewidth} >{\centering\arraybackslash}p{0.13\linewidth} >{\raggedright\arraybackslash}X@{}}
    \toprule
    Experiment & Configuration/protocol & Mean $\kappa$ & Interpretation \\
    \midrule
    Haryana zero-shot & Punjab-trained RF/very-strict/A; thresholds 0.30, 0.40, 0.30; no Haryana fitting or calibration & 0.641 & Promising neighboring-state transfer \\
    Punjab 2023 & Same seasonal framework & 0.662 & Stronger year-specific performance \\
    Punjab 2024 & Primary experiment & 0.393 & Weaker year-specific performance \\
    S1 + S2 fusion & Set E minus optical Set A & $-0.015\,\Delta\kappa$ & Simple concatenation did not improve performance \\
    \bottomrule
  \end{tabularx}
  \caption{Robustness and transfer summary. Cross-year reporting is limited to Punjab 2023 and 2024.}
  \label{tab:robustness}
\end{table}

\FloatBarrier

\section{Discussion}

\subsection{Interpreting Recipe-Specific Pseudo-Label Agreement}

The increase in agreement from moderate to very-strict recipe-specific pseudo-labels is the paper's central observed pattern. Tighter fire proximity, stronger spectral gates, and harder negative selection changed both the supervision and the evaluation population. Higher agreement may therefore reflect a more separable classification task as well as changes in training supervision. Because true label purity was not directly measured and the recipes were not evaluated against a common independent reference, the results do not establish improved label correctness or burned-area accuracy.

The near-equivalence of MLP, RF, XGBoost, and LightGBM under the strongest recipe also has an operational implication. Flexible classifiers cannot recover information that is absent or contradictory in the weak labels. Under the very-strict recipe, several nonlinear families achieved similar agreement with held-out pseudo-labels. The very-strict recipe-rule result reinforces this point: substantial discrimination was already encoded by the proxy rules. The role of learning was therefore to generalize beyond hard rule boundaries, integrate multiple change features, and produce a continuous statewide score rather than to create the signal from scratch. The feature-set ablations are also physically coherent with post-harvest burning: the reduction after removing SWIR-related predictors and the competitiveness of delta-only features are consistent with abrupt changes in vegetation condition, surface or canopy moisture, and shortwave-infrared response. These mechanisms are supportive rather than burn-specific, because harvest, tillage, moisture change, and atmospheric effects can produce overlapping spectral changes.

\subsection{Why External Calibration Was Necessary and What Scale It Supports}

The difference between the internal threshold of 0.30 and deployment threshold of 0.60 shows why weak-label performance cannot be transferred directly to mapping. Nested cross-validation selected a cutoff that best separated weak positives and negatives in held-out districts. Those classes were constructed through selective spatial and spectral rules and did not reproduce the prevalence or uncertainty of all cropland pixels. Statewide inference therefore required a second threshold-selection problem tied to mapped area and district patterns.

External calibration improved the plausibility of aggregate extent and district ranking, but it did not convert MODIS into ground truth. MCD64A1 has coarser spatial support and its own omission and commission characteristics, especially for small agricultural burns. District $R^2$, area ratio, and fire-count concordance are consequently more defensible indicators of regional burden than exact pixel overlap. The calibrated product is suited to district comparison, hotspot screening, and prioritization. It is not yet a reliable source of exact field boundaries, individual burn scars, or event timing.

\subsection{Transferability, Circularity, and Limitations}

Haryana transfer suggests that the selected optical-change representation was not restricted to the Punjab district partitions used for model development. However, the result is geographically narrow: Haryana is adjacent and shares broadly similar crop calendars and agricultural landscapes. The temporal contrast is more cautionary. A model that transfers across a neighboring state within a comparable seasonal context may still vary substantially across years because atmospheric conditions, residue management, crop phenology, and proxy-label coverage change.

Circularity was reduced but not eliminated. Excluding NBR and dNBR prevented the classifier from directly reusing the main label threshold, and no-SWIR ablation tested dependence on related spectral information. Yet the remaining optical predictors are correlated with the same physical burn process and with label-construction variables. This is scientifically appropriate for prediction, but it means internal weak-label metrics cannot be interpreted as independent pixel accuracy.

The optical windows do not bracket every fire in the seasonal reference period. Early burns may affect the earlier composite, while late-season scars are not observed by the later composite, which ends on 4 November. VIIRS context nevertheless extends through 29 November, and MODIS comprises the October and November monthly products. This temporal mismatch limits interpretation of agreement with the seasonal references.

The main limitations are the absence of an independent human-scored pixel-level reference dataset, heuristic weak labels, coarse calibration support, structured district residuals, weak 2023--2024 stability, and a negative result for only one simple radar-fusion design. Future work should prioritize independent field or visual reference sampling, year-adaptive label and threshold strategies, explicit uncertainty products, and structured temporal or multimodal fusion. Until then, the appropriate claim is calibrated regional monitoring under label scarcity, not operationally proven scar delineation.

\section{Conclusion}

This study developed a weakly supervised framework for post-harvest burned-cropland mapping that separates label construction, district-held-out learning, and map-level threshold calibration. The most selective pseudo-label recipe achieved the highest agreement with its own held-out pseudo-labels; this does not establish improved label correctness or independent burned-area accuracy. MLP/very-strict/Set A was the nominal best trained configuration (kappa 0.395), but RF, XGBoost, and LightGBM were practically tied within the very-strict recipe.

For production, RF/very-strict/Set A was retained because it could be deployed natively in Google Earth Engine with negligible and statistically non-significant loss relative to MLP. External deployment-threshold calibration was necessary: a cutoff of 0.60 yielded district MODIS $R^2$ of 0.636, an area ratio of 1.005, and fire-count Spearman correlation of 0.779, while pixel-level MODIS agreement remained modest. The final evidential scale is therefore district burden and hotspot screening rather than exact scar boundaries.

Early burns may affect the earlier optical composite, and late-season scars are not observed by the later composite despite the broader seasonal fire context. Zero-shot transfer to Haryana was promising, but Punjab performance differed sharply between 2023 and 2024, and simple Sentinel-1/Sentinel-2 concatenation did not improve the optical baseline. The framework therefore provides a defensible route to regional burned-cropland monitoring where labels are scarce, provided that pseudo-label recipes are treated as primary model components, deployment thresholds are calibrated against external aggregate evidence, and temporal and pixel-level limitations remain explicit. Operational use should focus on district-scale burden assessment, hotspot ranking, and screening rather than field-boundary or temporally invariant annual mapping. Structured district residuals and year-to-year variation should be reported alongside mapped totals, and independent human-scored reference sampling remains necessary before exact scar delineation can be claimed. The central methodological implication is that label design, spatially honest evaluation, and deployment calibration must be treated as a single evidence chain rather than as separate post-processing choices.

\begingroup
\footnotesize
\setlength{\bibsep}{1pt}
\bibliographystyle{plainnat}
\bibliography{references}
\endgroup

\end{document}